\documentclass[sigconf]{acmart}

\usepackage{booktabs}
\usepackage{graphicx}
\usepackage{grffile}   
\usepackage{amsmath}
\usepackage{multirow}
\usepackage{verbatim}
\usepackage{placeins}
\usepackage{balance}
\usepackage[ruled,linesnumbered,lined]{algorithm2e}

\SetKwInOut{KwInput}{Input}
\SetKwInOut{KwOutput}{Output}
\SetKwFunction{TermPresence}{TermPresenceScore}
\SetKwFunction{TermPass}{TermPasses}

\graphicspath{{image/}}
\definecolor{physred}{RGB}{170,45,45}
\definecolor{affblue}{RGB}{35,85,160}
\newcommand{\physterm}[1]{\textbf{\textcolor{physred}{#1}}}
\newcommand{\affterm}[1]{\textbf{\textcolor{affblue}{#1}}}
\makeatletter
\def\@subsubsecfont{\bfseries\section@raggedright}
\def\@parfont{\bfseries\itshape}
\makeatother

\setcopyright{acmlicensed}
\copyrightyear{2026}
\acmYear{2026}
\acmDOI{XXXXXXX.XXXXXXX}
\acmConference[ACM SIGSPATIAL '26]{34th ACM International Conference on
Advances in Geographic Information Systems}{November 3--6, 2026}{Riverside, CA, USA}
\acmBooktitle{Proceedings of the 34th ACM International Conference on
Advances in Geographic Information Systems (ACM SIGSPATIAL '26),
November 3--6, 2026, Riverside, CA, USA}
\acmISBN{978-X-XXXX-XXXX-X/26/11}

\author{Ziqi Cui}
\authornote{Both authors contributed equally to this work.}
\authornote{Corresponding author.}
\affiliation{%
  \institution{University of British Columbia}
  \city{Vancouver}
  \country{Canada}}
\email{ziqicui@student.ubc.ca}

\author{Shangyu Lou}
\authornotemark[1]
\affiliation{%
  \institution{University of California, Santa Barbara \& San Diego State University}
  \city{California}
  \country{USA}}
\email{shangyulou@ucsb.edu}

\title[PlaceSeek]{PlaceSeek: Human-Centered Geospatial Retrieval of Urban Outdoor Places via Semantic Grounding and Affective Alignment}

\begin{abstract}
People search for urban outdoor places not only by category or function,
but also by what activities a place can support and how it is perceived.
Existing geospatial retrieval remains largely POI-centric and
metadata-driven, making it difficult to satisfy open-ended, affective, or
activity-oriented needs. We present \textbf{PlaceSeek}, a human-centered
outdoor place retrieval framework that maps natural-language queries to
geolocated street-view imagery. PlaceSeek introduces an intent-aware
retrieval mechanism that decomposes user queries into functional and
affective sub-intents. A Semantic Grounding Module verifies whether
candidate street-view results contain the physical evidence needed to
support the intended activity, while an Affective Alignment Module
re-ranks physically valid candidates using a LoRA-adapted vision-language
model trained on human urban perception judgments. We evaluate PlaceSeek
on 31,956 street-view locations in Milan across 10 natural-language
queries annotated by five human evaluators. PlaceSeek achieves 88.0\%
Precision@5, a mean match score of 3.39/4.0, and 0.920 nDCG@5,
outperforming CLIP, fine-tuned CLIP, SigLIP, and a VQA-based baseline.
Ablation results show that physical grounding is essential for retrieval
validity, while affective alignment improves ranking quality among
physically valid candidates. These findings highlight that complex urban
spatial queries require modeling both verifiable visual evidence and
human perceptual preferences. PlaceSeek provides a potential framework
for human-centered next-generation geospatial retrieval systems.
\end{abstract}



\begin{document}
\maketitle

\section{Introduction}
\label{sec:intro}

Understanding and satisfying people's subjective needs in urban
environments is central to human-centered spatial decision-making~\cite{cui2025syncperception,ewing2009measuring}.
Urban search is commonly framed as the retrieval of
known categories: restaurants, parks, and other points of
interest (POIs)~\cite{purves2018geographic}. This paradigm works well when a user's need can be
mapped to a structured category and when the target place is supported
by rich metadata. Yet people's daily demands for urban space are often
more complex. A visitor in an unfamiliar city may not simply search for
a park or a cafe, but for ``a comfortable green space where I can sit
and relax'' or ``a shaded spot with public seating''. Such queries
combine activity support, physical facilities, and affective
expectations. They also extend beyond named POIs to include informal
outdoor places such as streetsides, plazas, pocket parks, and
waterfront edges.

\begin{figure}[htbp]
  \centering
  \includegraphics[width=\linewidth]{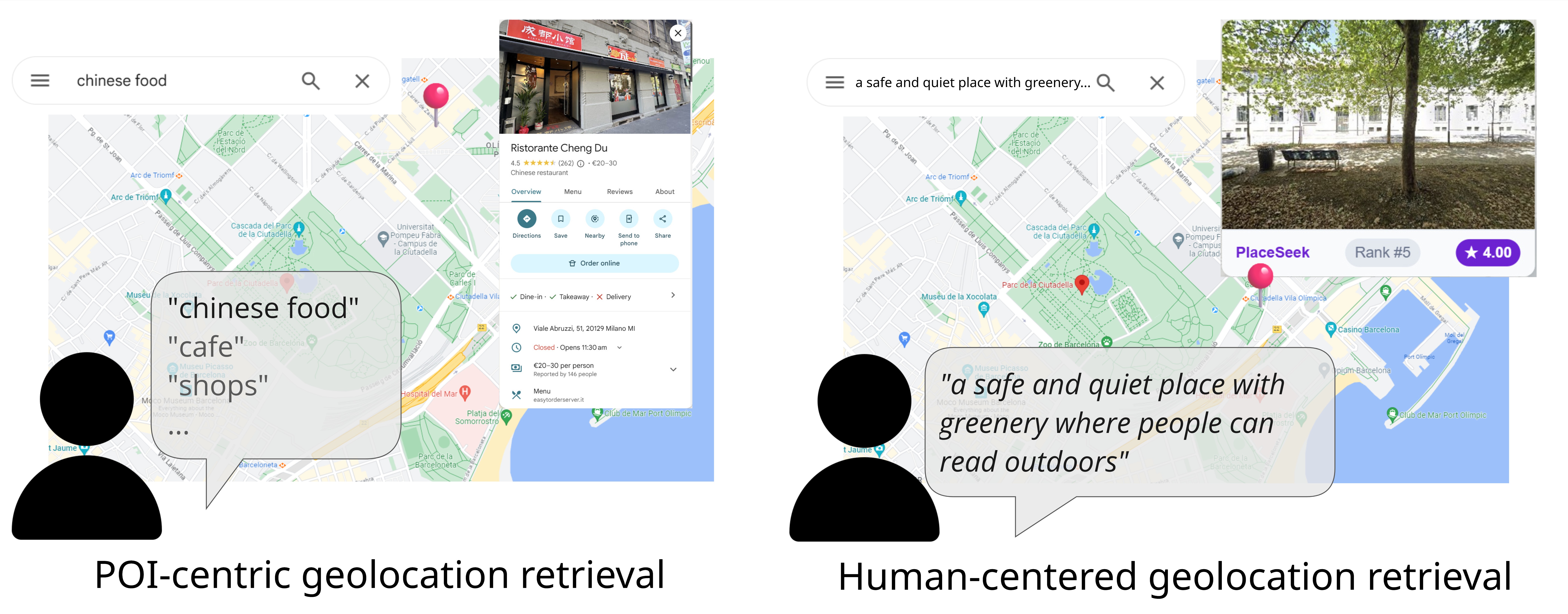}
  \caption{From POI-centric search to human-centered geolocation retrieval.}
  \label{fig:motivation}
\end{figure}

These needs are difficult to address using structured databases alone.
Outdoor open places usually do not have names, categories, or detailed
descriptions, and existing structured retrieval systems are limited in their ability
to interpret the abstract activity- and affect-related needs expressed
in natural language~\cite{purves2018geographic,tang2025uguiderag,khellaf2025spot}.
Figure~\ref{fig:motivation} illustrates
this shift from POI-centric retrieval toward human-centered geolocation
retrieval: rather than asking only where a known category is located,
users increasingly ask what kind of urban place can support a desired
activity and experience.

Street-view imagery (SVI) provides a promising basis for this problem
because of its extensive coverage and abundant visual information
\cite{li2015assessing,yao2017sensing}.
Recent vision-language models, such as CLIP and its variants, further
enable image retrieval from natural-language prompts without
task-specific labels~\cite{radford2021clip}. However, retrieval based
on global text-image similarity still has two fundamental limitations.
First, CLIP-style models are not explicitly designed to model human
affective preferences toward urban scenes, making it difficult to
reliably capture qualities such as safety, comfort, or romantic
atmosphere~\cite{salesses2013collaborative,ordonez2014urban,dubey2016placepulse}.
Second, retrieved results often lack support from concrete
physical evidence: a model may assign a high score to an image because
it broadly matches a dominant visual concept (e.g., greenery), while
failing to verify whether a smaller but necessary element (e.g., bench)
is actually present.

How can we retrieve urban outdoor places from abstract natural-language
descriptions of intended human activities and experiences? Based on the
idea that urban places support human behavior through both physical
affordances and perceived qualities~\cite{gibson1979ecological,ewing2009measuring,cui2025spatiallayout},
we argue that effective retrieval
requires two forms of evidence. First, concrete physical requirements
(e.g., benches, greenery, lighting, sidewalks) must be grounded in
visible street-view content. Second, retrieved places must align with
affective qualities that shape whether a place feels suitable for the
intended activity, such as safety, quietness, and comfort. To
operationalize this idea and address these limitations of geospatial retrieval, 
we develop \textit{PlaceSeek}, an intent-aware geospatial retrieval framework 
for finding urban outdoor places from natural-language queries based on SVIs.

This paper makes the following contributions:
\begin{itemize}
  \item We identify and formulate human-centered urban outdoor place
    retrieval as an underexplored geospatial retrieval problem. Unlike
    POI-centric search, this task targets unnamed or weakly indexed
    outdoor places from natural-language descriptions, and we propose
    \textit{PlaceSeek} as a solution framework for this setting.
  \item We design a Semantic Grounding Module (SGM) to address the
    limitation of global text-image matching, where visually plausible
    results may lack the required physical evidence. SGM introduces a
    coarse-to-fine physical evidence validation procedure that improves
    the success rate of top-$k$ retrieved SVIs in satisfying query-required elements.
  \item We develop an Affective Alignment Module (AAM) that improves
    human-perception alignment in text-to-street-view retrieval. AAM
    fine-tunes OpenCLIP with LoRA under supervision from real human urban
    perception judgments, enabling retrieved scenes to better match
    affective or experiential needs expressed in natural language.
  \item We conduct a human-annotated empirical evaluation on Milan
    SVIs across diverse natural-language queries.
    Comparisons with baseline methods, together with ablation analysis, demonstrate
    the effectiveness of PlaceSeek and the complementary roles of
    physical grounding and affective alignment.
\end{itemize}

\section{Related Work}
\label{sec:related}

\subsection{Human-Centered Geospatial Retrieval}
Traditional geospatial retrieval primarily targets structured spatial entities, 
such as POIs, roads, or buildings, that are represented through explicit 
categories, attributes, and spatial relations~\cite{purves2018geographic}. 
While effective for known and indexable objects, 
this paradigm is less suited to everyday spatial needs that are subjective, 
open-ended, and difficult to predefine~\cite{purves2018geographic,goodchild2007citizens}. Recent LLM- and RAG-based methods 
provide new opportunities for more human-centered geospatial retrieval. For example,
Spatial-RAG~\cite{yu2025spatialrag} integrates spatial databases with
LLMs for geospatial question answering and spatial reasoning;
UGuideRAG~\cite{tang2025uguiderag} uses intent-enhanced RAG and
user-generated content for personalized urban tourism recommendation;
SemaSK~\cite{zhang2025semask} uses LLMs to improve semantic matching in
spatial keyword queries over geo-textual objects such as
POIs~\cite{yelp2023}; and
SPOT~\cite{khellaf2025spot} converts natural-language scene descriptions
into structured OpenStreetMap object searches. These studies show that
natural-language interfaces can make geospatial search more flexible by
helping systems interpret user intent and connect it with spatial data.

However, existing methods still mainly retrieve POIs, structured GIS
objects or predefined map entities. They do not
directly address the retrieval of unnamed or weakly indexed outdoor
urban spaces, such as pocket parks, street corners, or informal resting
areas. As a result, users' needs for open outdoor places that are not
explicitly represented in spatial databases remain difficult to satisfy.

\subsection{Vision-Language Models for Open-Vocabulary Image Retrieval}
Vision-language models provide an important foundation for
open-vocabulary image retrieval. Dual-encoder models such as CLIP
\cite{radford2021clip}, OpenCLIP~\cite{cherti2023reproducible},
SigLIP~\cite{zhai2023sigmoid}, and related variants encode images and
text into a shared embedding space, enabling efficient retrieval over
large image collections through precomputed image embeddings and
text-image similarity. These models have become a major paradigm for
natural-language image retrieval and provide a scalable technical
pathway for SVI-based geospatial retrieval
\cite{li2015assessing,biljecki2021streetview}.

However, prior studies and practical applications have shown that
CLIP-style models remain limited in fine-grained image-text matching.
First, CLIP returns image-level global similarity scores and does not
provide explicit localized visual evidence~\cite{zhou2022dense}.
Therefore, it cannot guarantee that the specific physical element
required by a query is actually present. Open-vocabulary grounding and detection models, such
as GroundingDINO \cite{liu2023groundingdino}, can partially address
this issue by localizing candidate objects or regions from text
prompts. Yet in complex street-view scenes, these models may still be
affected by small object scale, occlusion, viewpoint variation, and
visually similar urban structures.

A second limitation is that CLIP-style models do not explicitly model
human affective and perceptual preferences toward urban scenes
\cite{salesses2013collaborative,dubey2016placepulse,selim2022urban}. As a result,
subjective affective queries, such as those involving safety,
quietness, comfort, or romantic atmosphere, are difficult to handle
reliably using generic CLIP similarity alone. In contrast, multimodal
large language models and visual question answering models
\cite{li2023blip2} can perform more fine-grained semantic judgment from
image content and natural-language instructions, making them promising
for subjective scene assessment. However, these models usually require
image-by-image inference, which is computationally expensive and
difficult to apply directly to city-scale street-view retrieval. Therefore, it remains challenging to maintain the efficiency of
dual-encoder text-image retrieval over large-scale multimodal
geospatial data while improving the top-$k$ match quality of retrieved
results in terms of both physical evidence and perceptual intent.

\section{Problem Formulation}
\label{sec:problem}

We formulate PlaceSeek as an intent-aware outdoor place retrieval problem
over city-scale street-view imagery (SVI). Given a user-provided
natural-language query, the task is to return a ranked list of geolocated
street-view results that provide visual evidence for candidate urban
places. Unlike conventional POI search, the target is not a named category
or establishment, but a place that satisfies the user's intended activity,
required physical evidence, and affective or experiential expectations.

\paragraph{Inputs and Outputs.}
The primary input is a user-provided natural-language query~$q$
describing a desired outdoor place. $q$ may explicitly or implicitly
contain a mixture of concrete visual requirements, affective
preferences, and intended activities. For example, a query such as
``a safe, quiet place with lots of greenery where people can read
outdoors'' encodes physical evidence and affordances
(\emph{greenery}, \emph{benches}), perceptual qualities
(\emph{safe}, \emph{quiet}), and activity intent (\emph{reading}).
The search space is a SVI database
$\mathcal{D}=\{(I_i, g_i, \theta_i)\}_{i=1}^{N}$, where each
image~$I_i$ is associated with a geographic location~$g_i$
(latitude/longitude) and a viewing direction~$\theta_i$. Given $q$
and $\mathcal{D}$, the system returns a ranked list of top-$k$
geolocated street-view results:
\[
  R_k(q)=\big[(I_{r_1}, g_{r_1}, \theta_{r_1}), \ldots,
  (I_{r_k}, g_{r_k}, \theta_{r_k})\big].
\]

\paragraph{Ranking Criteria.}
We seek a ranking function~$s(q,I_i)$ that orders images by three
complementary criteria: semantic relevance to the overall query,
grounding in explicit physical evidence (visible objects or
affordances required by the query), and alignment with the user's
affective and perceptual needs. Conceptually:
\[
  s(q,I_i)=
  \lambda_s s_{\mathrm{sem}}(q,I_i)
  + \lambda_g s_{\mathrm{ground}}(q,I_i)
  + \lambda_a s_{\mathrm{aff}}(q,I_i).
\]
In practice, PlaceSeek implements this as a staged pipeline: semantic
retrieval narrows the candidate set, physical grounding verifies
necessary visual elements, and affective alignment re-ranks the
verified candidates.

\paragraph{Core Challenges.}
This formulation introduces three challenges. First, user intents are
often underspecified and compositional: ``reading outdoors'' may imply
seating; ``walkable historic street'' may imply pedestrian space and
heritage architectural evidence. Second, physical validity cannot be
guaranteed by global visual similarity: CLIP-style models may miss
required elements or confuse visually similar objects such as benches,
railings, and curbs. Third, perceptual terms such as ``quiet'',
``romantic'', or ``safe'' are not reducible to object categories and
are difficult to capture with generic image-text embeddings alone.
These challenges motivate the three-component design of PlaceSeek.

\section{The PlaceSeek Framework}
\label{sec:system}

\begin{figure*}[t]
  \centering
  \includegraphics[width=\textwidth]{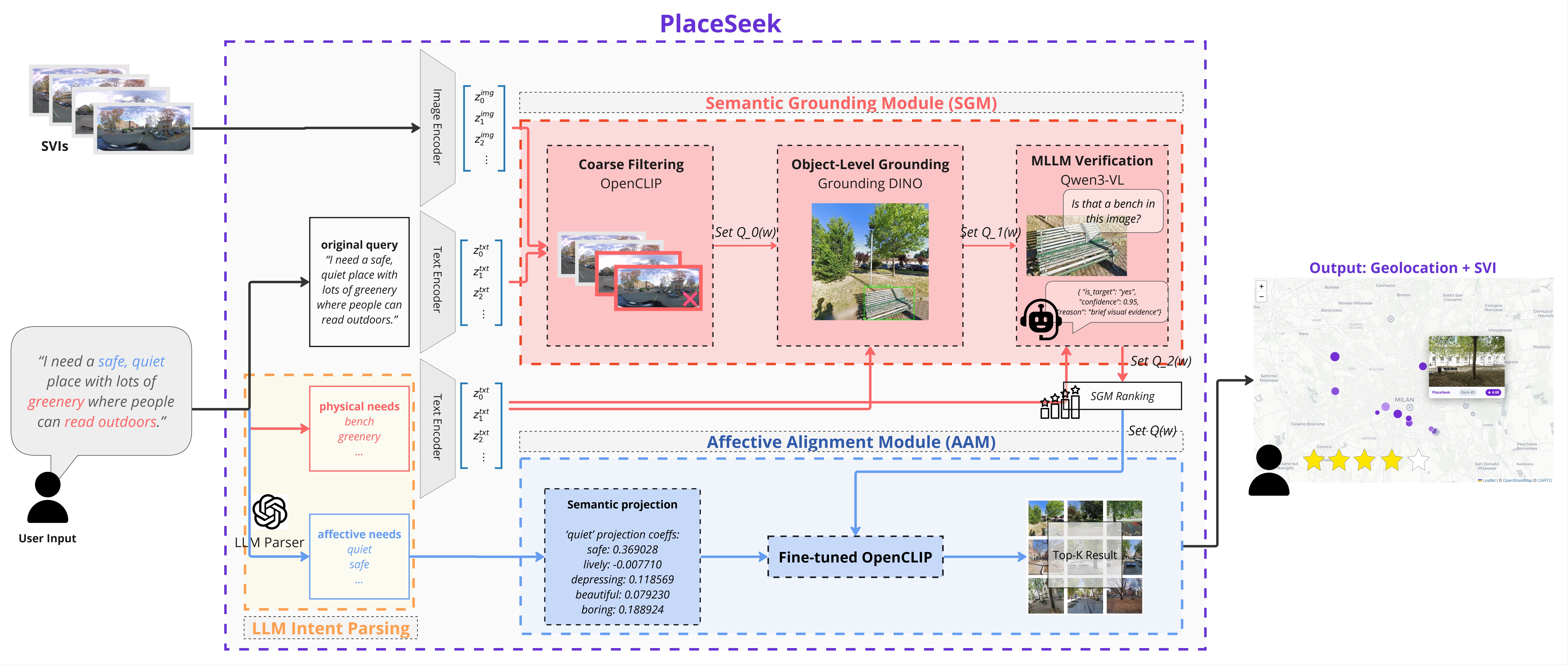}
  \caption{PlaceSeek pipeline. A user natural-language query is
    decomposed by an LLM parser into physical anchor words and affective need
    terms. The Semantic Grounding Module (SGM) performs coarse-to-fine
    image retrieval grounded in physical evidence; the Affective
    Alignment Module (AAM) then re-ranks the candidate set by
    perceptual alignment.}
  \label{fig:pipeline}
\end{figure*}

Figure~\ref{fig:pipeline} illustrates the overall PlaceSeek framework.
Given a user-provided natural-language query, the system proceeds through
three coordinated stages:

\begin{enumerate}
  \item \textbf{Intent Parsing.} An LLM decomposes the query into
    physical evidence requirements and affective or experiential
    preferences.
  \item \textbf{Semantic Grounding Module (SGM).} The SGM uses the
    physical requirements to retrieve and verify street-view candidates
    that contain the required visual evidence, producing a physically
    verified candidate set~$\mathcal{Q}$.
  \item \textbf{Affective Alignment Module (AAM).} The AAM re-ranks
    candidates in~$\mathcal{Q}$ according to their perceptual alignment
    with the user's affective or experiential expectations.
\end{enumerate}

The final output is a ranked list of top-$k$ geolocated outdoor place
results that best match the input query and can be inspected on a map.

\subsection{Intent Parsing}
\label{sec:intent-parsing}

The first stage converts an unstructured natural-language query into a
structured retrieval specification. We use ChatGPT-4o as an intent
parser~\cite{openai2024gpt4o} with a task-specific prompt that instructs the model to separate
visually verifiable physical evidence from affective, perceptual, and
experiential preferences. The parser output schema is summarized in
Appendix~\ref{app:prompts}.

Given a user query~$q$, the parser returns four fields:
\[
  \Pi(q)=\{\mathcal{P}, \mathcal{A}, \mathcal{U}, \mathcal{C}\},
\]
where $\mathcal{P}$ is the set of physical evidence requirements,
$\mathcal{A}$ is the set of affective or perceptual preferences,
$\mathcal{U}$ records intended activities, and $\mathcal{C}$ records
constraints or tensions in the query. Each parsed term is associated with
a source label, a requiredness level, and a short reasoning chain.

The prompt performs three parsing operations. First, it extracts explicit
physical evidence requirements from the query, including visible objects,
spatial affordances, or scene conditions such as greenery, benches,
statues, tram tracks, waterfronts, murals, or tall buildings. Second, it
applies activity-support inference when the query describes an intended
activity: for example, ``reading outdoors'' implies a need for seating or
a bench even when the word ``bench'' is not stated. Third, it identifies
affective or experiential preferences, including both explicitly stated
qualities such as ``safe'', ``quiet'', and ``lively'', and cautiously
inferred expectations that are strongly implied by the query context,
such as interpreting a desired date place as requiring a pleasant or
romantic atmosphere.

The output of this stage is a structured set of decomposed intent terms.
Downstream modules then consume these terms for physical grounding and
affective alignment. Table~\ref{tab:intent-example} shows an example decomposition for query
AO2. The example illustrates that the parser performs more than keyword
extraction: it preserves explicit visual evidence, infers
activity-supporting affordances, and separates subjective needs into
affective terms for downstream alignment. Appendix~\ref{app:intent-outputs} summarizes the generated physical and
affective intent outputs for all ten evaluation queries.

\begin{table}[t]
  \caption{Example intent decomposition for query AO2. The parser
    converts an activity-oriented request into decomposed physical and
    affective intent terms.}
  \label{tab:intent-example}
  \centering
  \small
  \setlength{\tabcolsep}{4pt}
  \begin{tabular}{p{0.25\linewidth}p{0.67\linewidth}}
    \toprule
    Component & Parsed output \\
    \midrule
    Query &
    ``A \affterm{safe}, \affterm{quiet} place with lots of
    \physterm{greenery} where people can \physterm{read outdoors}.'' \\
    Physical evidence terms &
    \physterm{greenery} (must; explicit; vegetation);
    \physterm{bench / public seating} (must; inferred from activity;
    affordance). \\
    Inference rule &
    \physterm{Reading outdoors} implies a sittable affordance;
    \physterm{benches or public seating} provide visible street-view
    evidence for outdoor reading. \\
    Affective terms &
    \affterm{safe} (positive; explicit; safety);
    \affterm{quiet} (positive; explicit; quietness). \\
    \bottomrule
  \end{tabular}
\end{table}

\subsection{Semantic Grounding}
\label{sec:sgm}

CLIP-style dual-encoder vision-language models provide an efficient mechanism for 
large-scale text-to-image retrieval~\cite{radford2021clip,cherti2023reproducible}. Through a dual-encoder architecture, 
images and text prompts are embedded into a shared vector space, 
allowing retrieval to be performed by nearest-neighbor search over 
precomputed street-view image embeddings. This dual-encoder design is
well suited for city-scale SVI retrieval. However, global image-text 
similarity does not guarantee the presence of specific physical evidence. 
In street-view imagery, we observed that CLIP often retrieves visually 
similar but semantically incorrect structures for object-level queries. 
For example, several top-ranked results for ``bench'' corresponded to 
railings or curbs rather than actual benches. The Semantic Grounding Module (SGM) 
therefore uses CLIP for coarse filtering, followed by object-level grounding 
and MLLM verification to ensure that retrieved candidates contain the required physical evidence.

\subsubsection{Coarse Filtering with OpenCLIP}

We encode all SVIs in the database using OpenCLIP ViT-L/14~\cite{cherti2023reproducible}. For each
physical evidence term~$w$, we construct a prompt bank~$B_w$, consisting
of short visual paraphrases of the target concept (e.g., ``bench'',
``park bench'', ``street bench'', ``outdoor bench'', and ``public
bench''). This reduces sensitivity to a single prompt wording and
improves recall during the initial retrieval stage. Each prompt is
encoded with the text encoder, and cosine similarity is computed against
all precomputed street-view image embeddings. For an image~$I_i$ and
prompt bank~$B_w$, the coarse score is:
\[
  s_{\mathrm{clip}}(I_i,w)
  = \operatorname{meanTop2}_{b \in B_w}
  \cos \big(f_I(I_i), f_T(b)\big),
\]
where $f_I$ and $f_T$ are the image and text encoders. We use the mean of
the top two prompt similarities to retain the strongest textual matches
while reducing the influence of weaker prompt variants.

Because our task is top-$k$ place retrieval rather than exhaustive
detection of all valid images, this coarse stage is designed to preserve
high-probability candidates. Images in the top 1\% of the resulting score
distribution are retained as the initial candidate set~$\mathcal{Q}_0(w)$,
balancing broad recall with the computational cost of the subsequent
grounding and verification stages.

\subsubsection{Object-Level Grounding with GroundingDINO}

GroundingDINO~\cite{liu2023groundingdino} is an open-vocabulary object
grounding model that localizes image regions corresponding to a given
text prompt. We apply it to each image in~$\mathcal{Q}_0(w)$ using the
corresponding physical evidence term~$w$ as the prompt. Unlike CLIP,
which provides only a global image-level similarity score, GroundingDINO
produces localized bounding boxes and confidence scores for candidate
visual evidence. For each image, we use the maximum bounding-box
confidence as the object-level grounding score and re-rank the candidates
accordingly, yielding~$\mathcal{Q}_1(w)$.

This stage improves physical specificity by filtering out many globally
similar but physically irrelevant images. However, detector confidence
alone is not sufficient for final verification: GroundingDINO may still
confuse visually similar structures, such as metal railings and benches,
or detect regions that match the object category but are not usable for
the intended activity. Therefore, the grounded candidates are further
checked by the MLLM verification stage.

\subsubsection{MLLM Verification with Qwen3-VL}

GroundingDINO does not fully determine whether the detected region is
semantically correct or functionally suitable. Qwen3-VL is a multimodal
large language model capable of interpreting visual inputs and producing
instruction-following textual outputs~\cite{bai2025qwen3vl}. For each image in
$\mathcal{Q}_1(w)$, the GroundingDINO-detected bounding region is cropped
and passed to Qwen3-VL using the verification prompt shown in
Appendix~\ref{app:prompts}. The MLLM returns a structured JSON output
containing a label (\texttt{yes}, \texttt{no}, or \texttt{uncertain}), a
confidence value, and a short visual rationale. The Qwen3-VL label serves as 
the primary semantic verification signal for physical evidence and 
contributes to the final rule-aware physical ranking. This produces a
verified per-term evidence set~$\mathcal{Q}_2(w)$.

\subsubsection{Physical Evidence Re-ranking}

A user query may contain multiple physical evidence terms, such as
\emph{bench} and \emph{greenery}. SGM first constructs evidence for each
term independently through OpenCLIP retrieval, GroundingDINO localization,
and Qwen3-VL verification. The verified per-term evidence is then merged
at the panorama/view level so that each candidate view can be evaluated
against the full set of physical requirements in the query. We denote the
merged, rule-aware physical candidate set as~$\mathcal{Q}$.

For each physical term~$w$ and candidate view~$v$, let
$\ell_w(v)\in\{\text{yes},\text{uncertain},\text{no},\text{missing}\}$
denote the normalized Qwen3-VL verification label, and let
$c_w(v)\in[0,1]$ denote its confidence. We first compute a detector-based
presence score from GroundingDINO evidence:
\[
  d_w(v)=
  \alpha_1 s_{\mathrm{box}}
  +\alpha_2 \hat{n}_{\mathrm{box}}
  +\alpha_3 \hat{a}_{\mathrm{box}},
  \quad
  \alpha_j \ge 0,\quad
  \sum_{j=1}^{3}\alpha_j=1,
\]
where $s_{\mathrm{box}}$ is the maximum box confidence,
$\hat{n}_{\mathrm{box}}$ is the normalized box count, and
$\hat{a}_{\mathrm{box}}$ is the normalized largest box area ratio. This
score captures whether the target evidence is detected confidently,
repeatedly, and at a visually meaningful scale.

The Qwen3-VL evidence score is defined as
\[
  q_w(v)=\phi(\ell_w(v))\cdot c_w(v),
\]
where $\phi(\text{yes})=1$, $\phi(\text{uncertain})=0.5$, and
$\phi(\text{no})=\phi(\text{missing})=0$. The term-level physical
presence score is then computed as:
\[
  p_w(v)=\eta q_w(v)+(1-\eta)d_w(v),
  \quad \eta>0.5.
\]
This fusion gives higher weight to Qwen3-VL because it provides the
primary semantic verification signal, while GroundingDINO is retained as
auxiliary localized evidence. Fixed weight values and normalization
details are provided in Appendix~\ref{app:ranking-details}.

To aggregate multiple physical evidence terms into a view-level physical
assessment, each term is assigned a rule label during LLM-based intent
parsing. Each physical term is treated as one of four types:
\emph{must-have}, \emph{more-better}, \emph{less-better}, or
\emph{not-exist}. These types define both physical validity and ranking
contribution. Must-have terms and not-exist terms define hard
constraints: candidates must satisfy all required evidence and must not
violate forbidden evidence. More-better and less-better terms define soft
preferences that affect ranking but do not by themselves determine
validity.

Formally, candidates must satisfy the hard physical gate:
\[
  \text{pass}_{\mathrm{hard}}(v)=
  \big(\text{\#must-pass}=\text{\#must-total}\big)
  \land \big(\text{\#forbid-violations}=0\big).
\]
Candidates with $\text{pass}_{\mathrm{hard}}(v)=0$ are removed. For the
remaining candidates, term-level scores~$p_w(v)$ are aggregated according
to their rule types to obtain a rule-aware physical score
$s_{\mathrm{phys}}(v)$. The valid candidates are then ranked by
$s_{\mathrm{phys}}(v)$ and normalized CLIP evidence, yielding the
physically verified candidate set~$\mathcal{Q}$.
Algorithm~\ref{alg:physical-rerank} summarizes the rule-aware reranking
procedure.

\begin{algorithm}[t]
  \small
  \caption{Physical Rule-Aware Reranking}
  \label{alg:physical-rerank}
  \KwInput{Per-term Qwen tables $\{T_w\}$; physical rules $\mathcal{R}$;
    CLIP scores $\{s_{\mathrm{clip}}(v)\}$}
  \KwOutput{Filtered, ranked candidate set $\mathcal{Q}$}
  Merge all $T_w$ by panorama/view key into one table $T$\;
  \ForEach{view $v \in T$ and rule term $w \in \mathcal{R}$}{
    Compute $p_w(v) \leftarrow \TermPresence(v,w)$\;
    Set $\text{pass}_w(v) \leftarrow \TermPass(v,w,p_w(v))$\;
  }
  \ForEach{view $v \in T$}{
    Compute $s_{\mathrm{must}}(v)$, $s_{\mathrm{more}}(v)$,
    $s_{\mathrm{less}}^{\mathrm{clean}}(v)$, and
    $s_{\mathrm{forbid}}^{\mathrm{clean}}(v)$\;
    Compute $s_{\mathrm{phys}}(v)$ using the active rule weights\;
    Set $\text{pass}_{\mathrm{hard}}(v)$ if all must-have terms pass and
    no forbidden term violates\;
  }
  Remove all views with $\text{pass}_{\mathrm{hard}}(v)=0$\;
  Sort remaining views by
  $(\text{\#must-pass}\downarrow,\;
  \text{\#forbid-violations}\uparrow,\;
  s_{\mathrm{phys}}\downarrow,\;
  \tilde{s}_{\mathrm{clip}}\downarrow)$\;
  Assign $\text{final\_rank}$ by sort order\;
\end{algorithm}

\subsection{Affective Alignment}
\label{sec:aam}

\subsubsection{Model Fine-Tuning via LoRA}
CLIP-style models do not explicitly model human affective preferences
toward urban scenes. To improve alignment with abstract affective
semantics in user queries, we fine-tune OpenCLIP using human perceptual
supervision. We use Place Pulse 2.0~\cite{dubey2016placepulse}, a
street-view dataset containing pairwise human judgments across six
perceptual dimensions: safer, livelier, more beautiful, wealthier, more
depressing, and more boring.

We fine-tune the image encoder of OpenCLIP ViT-L/14 using additive LoRA
adapters~\cite{hu2022lora} on the Place Pulse 2.0 training split. The
text encoder is frozen to preserve zero-shot textual retrieval
capability, while the image encoder is adapted toward human perceptual
judgments. We use a learning rate of $10^{-4}$ and a batch size of 60,
corresponding to 10 image pairs across the six perceptual dimensions.

We evaluate the adapted model using pairwise win rate on a held-out Place
Pulse 2.0 test set. Baselines include unmodified OpenCLIP ViT-B/32,
SigLIP SO400M, EVA-CLIP L/14~\cite{sun2023evaclip}, and OpenCLIP ViT-L/14.
Table~\ref{tab:aam} reports win rates across all six perceptual
dimensions. The fine-tuned model improves the macro-average win rate from
approximately 52\% for unmodified CLIP-style models to 65.7\%, achieving
a 7--12.6 percentage point gain over the baselines. This indicates that
generic image--text pretraining alone does not reliably encode human
perceptual judgments, while lightweight LoRA adaptation improves
affective alignment for urban street-view retrieval.

\begin{table*}[t]
  \caption{Pairwise win rates (\%) across six perceptual dimensions.
    Best per column in \textbf{bold}.}
  \label{tab:aam}
  \centering
  \setlength{\tabcolsep}{5pt}
  \begin{tabular}{lccccccr}
    \toprule
    Model & Liv. $\uparrow$ & Beauty $\uparrow$ & Boring $\uparrow$ & Depr. $\uparrow$ & Safe $\uparrow$ & Wealth $\uparrow$ & Avg. $\uparrow$ \\
    \midrule
    EVA-CLIP L/14          & 55.0 & 51.2 & 45.8 & 52.0 & 51.8 & 47.8 & 50.6 \\
    SigLIP SO400M          & 55.1 & 56.3 & 46.9 & 48.3 & 55.9 & 56.1 & 53.1 \\
    OpenCLIP B/32          & 57.6 & 53.0 & 45.9 & 47.2 & 55.6 & 56.3 & 52.6 \\
    OpenCLIP L/14          & 55.6 & 55.7 & 49.2 & 53.9 & 56.3 & 52.9 & 52.8 \\
    \textbf{OCLIP L/14 (ours)} & \textbf{64.6} & \textbf{68.6} & \textbf{60.7} & \textbf{65.9} & \textbf{65.4} & \textbf{68.6} & \textbf{65.7} \\
    \bottomrule
  \end{tabular}
\end{table*}

\subsubsection{Affective Prompt Projection}

Parsed affective terms~$a$ from the user query, such as
``quiet'' or ``romantic'', may not directly correspond
to one of the six Place Pulse dimensions. We therefore represent each
affective term using two complementary signals: a projected perceptual
score in the Place Pulse space and a direct prompt-bank similarity score.

Let $B_a$ denote the prompt bank for affective term~$a$, and let
\[
  \mathbf{u}_a=\operatorname{mean}_{b\in B_a} f_T(b)
\]
be its normalized mean text embedding. We also encode six axis prompts
corresponding to the Place Pulse dimensions:
\[
  \mathcal{X}=\{\text{safe},\text{lively},\text{beautiful},
  \text{wealthy},\text{depressing},\text{boring}\}.
\]
The affective term embedding~$\mathbf{u}_a$ is mapped to this
six-dimensional perceptual axis set to obtain a proxy affective
direction~$\hat{\mathbf{u}}_a$. This proxy direction provides a
human-perception-guided representation of the affective term within the
Place Pulse space.

For each candidate image~$I_i$, we compute an affective relevance score
by combining the proxy perceptual score with direct prompt-bank
similarity:
\[
  s_a(I_i)=
  \lambda_{\mathrm{proj}}
  \cos(f_I^{\mathrm{aff}}(I_i),\hat{\mathbf{u}}_a)
  +
  \lambda_{\mathrm{dir}}
  \operatorname{mean}_{b\in B_a}
  \cos(f_I^{\mathrm{aff}}(I_i),f_T(b)),
\]
where $f_I^{\mathrm{aff}}$ is the LoRA-adapted image encoder, and
$\lambda_{\mathrm{proj}}+\lambda_{\mathrm{dir}}=1$. The projected term
captures alignment with human perceptual dimensions, while the direct
prompt-bank term preserves the original semantic meaning of the query
expression.

For queries with multiple affective terms, we average term-level scores
to obtain the raw affective score:
\[
  s_{\mathrm{aff}}^{\mathrm{raw}}(v)=\frac{1}{|\mathcal{A}|}
  \sum_{a\in\mathcal{A}} s_a(v).
\]
This score is used in the final physical-affective re-ranking stage.

\subsubsection{Final Physical-Affective Re-ranking}
The final ranking is computed over the physically verified candidate
set~$\mathcal{Q}$. For each candidate view~$v$, we combine three signals:
the rule-aware physical score~$s_{\mathrm{phys}}(v)$, the normalized
affective score~$\tilde{s}_{\mathrm{aff}}(v)$, and the normalized CLIP
coarse-retrieval score~$\tilde{s}_{\mathrm{clip}}(v)$. The final score is
defined as:
\[
  s_{\mathrm{final}}(v)=
  \omega_{\mathrm{phys}}s_{\mathrm{phys}}(v)
  +\omega_{\mathrm{aff}}\tilde{s}_{\mathrm{aff}}(v)
  +\omega_{\mathrm{clip}}\tilde{s}_{\mathrm{clip}}(v),
\]
where $\omega_{\mathrm{phys}}+\omega_{\mathrm{aff}}+\omega_{\mathrm{clip}}=1$.
The weighting scheme follows the staged design of PlaceSeek: the physical
score carries the main evidence from the SGM, the affective score refines
the ordering using the AAM, and the CLIP score retains a weak global
semantic prior from the coarse retrieval stage. The fixed weight values
and normalization details are provided in
Appendix~\ref{app:ranking-details}.

Final ranking preserves the physical gate: candidates that fail required
physical evidence or violate forbidden evidence are not considered. The
remaining candidates are sorted by~$s_{\mathrm{final}}(v)$, with raw
affective score and normalized CLIP similarity used only as tie-breakers.
This produces the final ranked list~$R_k(q)$.
Algorithm~\ref{alg:affective-rerank} summarizes this step.

\begin{algorithm}[t]
  \small
  \caption{Physical-Affective Final Reranking}
  \label{alg:affective-rerank}
  \KwInput{Physically verified candidate set $\mathcal{Q}$ with
    $s_{\mathrm{phys}}(v)$ and $\tilde{s}_{\mathrm{clip}}(v)$;
    per-term affective scores $\{s_{\mathrm{aff},t}^{\mathrm{raw}}(v)\}$}
  \KwOutput{Final ranked list $R_k(q)$}
  \ForEach{view $v \in \mathcal{Q}$}{
    Set $s_{\mathrm{aff}}^{\mathrm{raw}}(v) \leftarrow
    \operatorname{mean}_t\, s_{\mathrm{aff},t}^{\mathrm{raw}}(v)$\;
    Set $\tilde{s}_{\mathrm{aff}}(v) \leftarrow
    \textsc{MinMaxNorm}(s_{\mathrm{aff}}^{\mathrm{raw}}(v))$\;
    Compute $s_{\mathrm{final}}(v) \leftarrow
    \omega_{\mathrm{phys}}s_{\mathrm{phys}}(v)
    +\omega_{\mathrm{aff}}\tilde{s}_{\mathrm{aff}}(v)
    +\omega_{\mathrm{clip}}\tilde{s}_{\mathrm{clip}}(v)$\;
  }
  Remove views that fail the physical gate\;
  Sort remaining views by
  $(s_{\mathrm{final}}\downarrow,\;
  s_{\mathrm{aff}}^{\mathrm{raw}}\downarrow,\;
  \tilde{s}_{\mathrm{clip}}\downarrow)$\;
  \Return top-$k$ views as $R_k(q)$\;
\end{algorithm}

\section{Experimental Setup}
\label{sec:eval}

This study evaluates PlaceSeek through an end-to-end retrieval task: given a
natural-language query, each method retrieves a ranked list of
SVIs, and the top-20 results are assessed by human
annotators for query match.

\paragraph{Study Area and Data.}
We use Milan, Italy as study area. Street-view imagery was collected
from Google Street View~\cite{anguelov2010google} and sampled at 100\,m
intervals along the road network at four viewing directions
(0\textdegree, 90\textdegree, 180\textdegree, 270\textdegree),
yielding 127,824 geo-referenced images across 31,956 locations.

\paragraph{Query Tasks.}
We design ten natural-language test queries to cover diverse outdoor
place intents involving activities, physical evidence, and affective or
experiential preferences. The queries are grouped into four complementary
types, as summarized in Table~\ref{tab:queries}. \emph{Activity-oriented}
queries describe intended uses and often require inferring supporting
spatial affordances. \emph{Object-oriented} queries emphasize concrete
visible elements whose presence can be checked in street-view imagery.
\emph{Perception-oriented} queries primarily express desired felt
qualities, such as comfort, safety, or wealth. \emph{Mixed-intent}
queries combine perceptual, spatial, and object-level requirements in a
single request.

\begin{table*}[t]
  \caption{The ten test queries used in the end-to-end evaluation,
    grouped by query type.}
  \label{tab:queries}
  \centering
  \setlength{\tabcolsep}{5pt}
  \begin{tabular}{llp{11cm}}
    \toprule
    ID & Category & Query Text \\
    \midrule
    AO1 & Activity-oriented & I want to find a walkable street surrounded by vintage-style historic buildings. \\
    AO2 & Activity-oriented & A safe and quiet place with greenery where people can read outdoors. \\
    AO3 & Activity-oriented & A tree-lined shaded path suitable for jogging. \\
    MI1 & Mixed-intent & I'm looking for a romantic public square with a visible statue or sculpture. \\
    MI2 & Mixed-intent & I'm looking for a lively waterside place in the city. \\
    MI3 & Mixed-intent & I want to find a modern-looking urban place with tall buildings around. \\
    OO1 & Object-oriented & A well-maintained street with artistic murals or graffiti. \\
    OO2 & Object-oriented & I'm looking for a tree-lined street with tram tracks running down the middle. \\
    PO1 & Perception-oriented & A relaxing and comfortable public square. \\
    PO2 & Perception-oriented & A wealthy and safe street. \\
    \bottomrule
  \end{tabular}
\end{table*}

\subsection{Comparison Methods}
We compare PlaceSeek with four baselines that represent different
strategies for text-to-street-view retrieval.

\begin{enumerate}
  \item \textbf{CLIP}: a generic vision-language retrieval baseline. We
    encode the full user query with OpenCLIP ViT-L/14 and rank all SVIs
    by query-image cosine similarity. This baseline tests how well
    standard global image--text similarity can handle human-centered
    outdoor place queries without any task-specific adaptation.

  \item \textbf{FT-CLIP}: an affectively adapted CLIP retrieval baseline.
    We use the LoRA fine-tuned OpenCLIP model from the AAM and apply the
    full user query directly. This baseline tests whether affective
    fine-tuning alone is sufficient to improve retrieval quality for
    complex place intents.

  \item \textbf{SigLIP}: a stronger zero-shot CLIP-style retrieval
    baseline. We encode the full query and rank SVIs using SigLIP
    SO400M, a vision-language model trained with a
    sigmoid loss objective~\cite{zhai2023sigmoid}. This baseline tests whether performance gains
    can be achieved simply by replacing OpenCLIP with a stronger
    general-purpose retrieval model.

  \item \textbf{VQA (Qwen3)}: an MLLM-based Visual Question Answering (VQA)
    baseline. MLLMs are strong at instruction-following visual
    understanding and can assess whether an image satisfies a complex
    natural-language description. However, image-by-image inference limits
    their applicability to large-scale text-image retrieval. We therefore
    first use OpenCLIP for coarse filtering to obtain a manageable
    candidate set, and then prompt Qwen3-VL to judge the degree of match
    between each candidate image and the full user query, together with a
    short supporting rationale. This baseline tests whether a strong VQA
    model can solve the retrieval task directly without explicit intent
    decomposition or object-level grounding.
\end{enumerate}
\subsection{Annotation Protocol and Ground-Truth}
\label{sec:annotation}

\paragraph{Annotators and ethics.}
End-to-end relevance labels were collected from five independent
annotators with backgrounds spanning urban design, GIS, and
non-specialist perspectives. All annotators participated voluntarily and were fully informed of the
task purpose; no personally identifiable information was collected.

\paragraph{Annotation task.}
For each method and query, the top-20 retrieved SVIs were assessed
independently by all five annotators. Each candidate was rated on a
4-point Likert scale for overall query match ($1 =$ not a match,
$2 =$ weak match, $3 =$ good match, and $4 =$ perfect match), as well
as for separate physical and affective match dimensions using the same
scale. Annotators viewed only the query text and the street-view image.

\paragraph{Ground-truth aggregation.}
We binarize overall match scores into non-match (1--2) and match (3--4),
then apply majority voting across the five annotators, requiring at least
three votes for the majority label. The final aggregated relevance score is
the mean of annotator scores within that majority bucket, reducing outlier
influence while preserving graded relevance. Candidates with aggregated
scores $\geq 3$ are counted as successful matches for Precision@$k$; mean
match and nDCG@$k$ use the same aggregated scores, with nDCG gain defined
as $\max(\text{score}-2,0)$.

\paragraph{Inter-annotator agreement.}
To assess label reliability, we report binary Fleiss' $\kappa$~\cite{fleiss1971measuring} on
903 annotated candidate--query pairs, treating scores $\leq 2$ as
non-match and scores $\geq 3$ as match. Table~\ref{tab:agreement}
summarizes agreement on overall, physical, and affective match.
Physical match is judged most consistently ($\kappa = 0.631$), while
affective match is more subjective ($\kappa = 0.430$). Overall match,
which directly defines retrieval relevance, achieves moderate agreement
($\kappa = 0.522$; unanimous binary agreement on 51.7\% of items).
Physical attributes are more reliably verifiable, while affective
qualities remain inherently subjective.

\begin{table}[t]
  \caption{Inter-annotator agreement on 903 candidate--query pairs
    ($n = 5$ annotators). Binary labels treat scores $\leq 2$ as
    non-match and $\geq 3$ as match.}
  \label{tab:agreement}
  \centering
  \small
  \setlength{\tabcolsep}{4pt}
  \begin{tabular}{lcc}
    \toprule
    Label dimension & Fleiss' $\kappa$ & Unanimous agreement (\%) \\
    \midrule
    Overall match   & 0.522 & 51.7 \\
    Physical match  & 0.631 & 64.1 \\
    Affective match & 0.430 & 44.4 \\
    \bottomrule
  \end{tabular}
\end{table}

\subsection{Evaluation Metrics}

We report three complementary metrics at cutoffs $k \in \{5,10,20\}$,
letting $r_i \in [1,4]$ denote the aggregated relevance score at rank~$i$
(Section~\ref{sec:annotation}).
\textbf{Mean match} is the average graded relevance
$\mathrm{MeanMatch@}k=\frac{1}{k}\sum_{i=1}^{k}r_i$.
\textbf{Precision@$k$} is the fraction of successful matches
($r_i\!\geq\!3$): $\mathrm{Precision@}k=\frac{1}{k}\sum_{i=1}^{k}\mathbb{I}(r_i\!\geq\!3)$.
\textbf{nDCG@$k$}~\cite{jarvelin2002cumulated} uses gain $g_i=\max(r_i-2,0)$,
so non-matches contribute zero gain:
\[
  \mathrm{nDCG@}k=\frac{\sum_{i=1}^{k}g_i/\log_2(i+1)}{\mathrm{IDCG@}k}.
\]
All metrics are averaged across the ten test queries.
\section{Results}
\label{sec:results}
\raggedbottom

\subsection{Main Results}
\begin{figure}[!t]
  \centering
  \includegraphics[width=\linewidth]{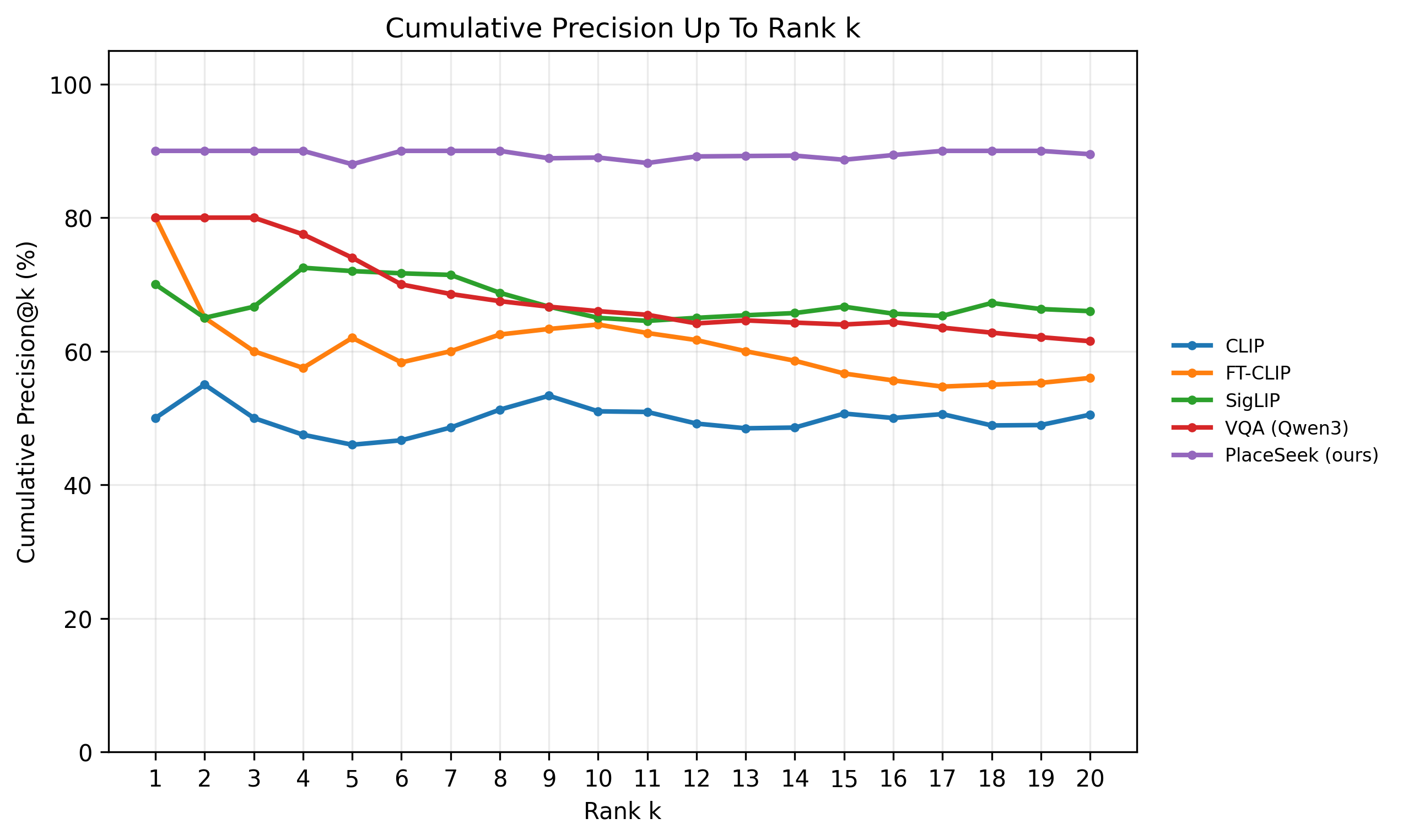}
  \caption{Cumulative precision ($\uparrow$) curves averaged over all 10 queries.
    PlaceSeek maintains the highest precision across the top-20 ranked
    results, while baselines degrade more rapidly.}
  \label{fig:cumprec}
\end{figure}

Table~\ref{tab:e2e} and Figure~\ref{fig:cumprec} summarize end-to-end
retrieval performance averaged over all ten test queries. PlaceSeek
achieves the best overall performance across all evaluated cutoffs, with
the strongest gains in graded relevance and rank-sensitive quality. At
top-5, PlaceSeek obtains a Precision@5 of 88.0\%, a mean match score of
3.39/4.0, and an nDCG@5 of 0.920. The strongest baseline varies by
metric: VQA (Qwen3) achieves the highest baseline Precision@5 (74.0\%)
and nDCG@5 (0.884), while SigLIP obtains a comparable mean match score
(2.92/4.0). However, PlaceSeek remains consistently ahead of both methods.

The advantage becomes more pronounced at larger cutoffs.
Precision@20 remains 89.5\% for PlaceSeek, compared with 66.0\% for
SigLIP, 61.5\% for VQA (Qwen3), and 50.5\% for CLIP.
Figure~\ref{fig:cumprec} shows the same trend across rank positions:
PlaceSeek maintains the highest cumulative precision through most of the
top-20 result list, while baseline methods degrade more rapidly. This
suggests that PlaceSeek improves not only early-rank retrieval, but also
the quality of the broader candidate set available for map-based place
exploration.

\begin{table*}[!t]
  \caption{End-to-end retrieval performance on 10 test queries (mean over
    all queries). Best result per column in \textbf{bold}.}
  \label{tab:e2e}
  \setlength{\tabcolsep}{5pt}
  \begin{tabular}{lrrrrrrrrr}
    \toprule
    & \multicolumn{3}{c}{Mean match (1--4) $\uparrow$}
    & \multicolumn{3}{c}{Precision@$k$ (\%) $\uparrow$}
    & \multicolumn{3}{c}{nDCG@$k$ $\uparrow$} \\
    \cmidrule(r){2-4}\cmidrule(r){5-7}\cmidrule(r){8-10}
    Method & @5 & @10 & @20 & @5 & @10 & @20 & @5 & @10 & @20 \\
    \midrule
    CLIP             & 2.43 & 2.49 & 2.47 & 46.0 & 51.0 & 50.5 & 0.650 & 0.668 & 0.695 \\
    FT-CLIP          & 2.75 & 2.81 & 2.61 & 62.0 & 64.0 & 56.0 & 0.774 & 0.743 & 0.736 \\
    SigLIP           & 2.92 & 2.78 & 2.79 & 72.0 & 65.0 & 66.0 & 0.825 & 0.825 & 0.816 \\
    VQA (Qwen3)      & 2.93 & 2.81 & 2.71 & 74.0 & 66.0 & 61.5 & 0.884 & 0.867 & 0.830 \\
    \textbf{PlaceSeek (ours)} & \textbf{3.39} & \textbf{3.36} & \textbf{3.37}
                              & \textbf{88.0} & \textbf{89.0} & \textbf{89.5}
                              & \textbf{0.920} & \textbf{0.917} & \textbf{0.910} \\
    \bottomrule
  \end{tabular}
\end{table*}

Table~\ref{tab:component-scores} decomposes the top-5 mean match score
into overall, physical, and affective dimensions. PlaceSeek achieves the
highest score in all three dimensions, with a physical match score of
3.63/4.0 and an affective match score of 3.41/4.0. Compared with the
strongest baseline in each dimension, PlaceSeek improves physical match by
0.29 points and affective match by 0.43 points. This indicates that the
overall improvement is not limited to a single aspect of the query, but
appears across both physical evidence matching and affective alignment.

\begin{table}[!ht]
  \caption{Top-5 mean match scores decomposed into overall, physical,
    and affective dimensions (1--4 scale). Best result per column in
    \textbf{bold}.}
  \label{tab:component-scores}
  \centering
  \small
  \setlength{\tabcolsep}{5pt}
  \begin{tabular}{lccc}
    \toprule
    Method & Overall & Physical & Affective \\
    \midrule
    CLIP             & 2.43 & 2.72 & 2.46 \\
    FT-CLIP          & 2.75 & 2.92 & 2.87 \\
    SigLIP           & 2.92 & 3.34 & 2.97 \\
    VQA (Qwen3)      & 2.93 & 3.09 & 2.93 \\
    \textbf{PlaceSeek (ours)}
                     & \textbf{3.39} & \textbf{3.63} & \textbf{3.41} \\
    \bottomrule
  \end{tabular}
\end{table}

\subsection{Ablation Study}
To quantify the contribution of each major component, we evaluate two
ablated variants of PlaceSeek:
\begin{itemize}
  \item \textbf{PlaceSeek w/o SGM}: AAM is applied directly to the full
    query embedding without physical evidence grounding, corresponding to
    affective re-ranking over raw OpenCLIP retrieval results.
  \item \textbf{PlaceSeek w/o AAM}: SGM is used without the final
    affective re-ranking step, corresponding to physical evidence
    grounding only.
\end{itemize}
Table~\ref{tab:ablation} shows that the two modules contribute in
different ways. Removing SGM causes the largest degradation:
Precision@5 drops from 88.0\% to 38.0\%, and Precision@20 drops from
89.5\% to 46.5\%. This indicates that AAM alone is insufficient to
guarantee the presence of required visual elements, particularly for
queries with explicit physical evidence requirements.

Removing AAM produces a more subtle pattern. The w/o AAM variant
matches the full model in Precision@5, but its nDCG@5 is lower. This
suggests that both methods retrieve the same proportion of successful
matches in the top-5, but the full PlaceSeek model places
higher-quality matches earlier in the ranking. At larger cutoffs, the
contribution of AAM becomes more evident: Precision@20 drops from
89.5\% in the full model to 74.5\% without AAM, while the full model
also maintains higher nDCG across all cutoffs. This suggests that AAM
primarily improves rank-sensitive ordering and candidate-set stability:
among physically valid candidates, it helps prioritize those whose
perceived atmosphere better matches the user's affective or experiential
intent.

Overall, the ablation results demonstrate the necessity of both physical
evidence grounding and affective alignment in PlaceSeek. SGM prevents
visually plausible but physically invalid scenes from entering the final
result set, while AAM refines the ordering within the grounded candidate
set to improve perceptual fit.

\begin{table}[!ht]
  \caption{Ablation study. Best result per column in \textbf{bold}.}
  \label{tab:ablation}
  \centering
  \small
  \setlength{\tabcolsep}{3pt}
  \begin{tabular}{lrrrrrr}
    \toprule
    & \multicolumn{2}{c}{@5} & \multicolumn{2}{c}{@10} & \multicolumn{2}{c}{@20} \\
    \cmidrule(r){2-3}\cmidrule(r){4-5}\cmidrule(r){6-7}
    Method & Prec. $\uparrow$ & nDCG $\uparrow$ & Prec. $\uparrow$ & nDCG $\uparrow$ & Prec. $\uparrow$ & nDCG $\uparrow$ \\
    \midrule
    w/o SGM & 38.0 & 0.466 & 43.0 & 0.507 & 46.5 & 0.591 \\
    w/o AAM & \textbf{88.0} & 0.914 & 83.0 & 0.906 & 74.5 & 0.894 \\
    \textbf{Full PlaceSeek} & \textbf{88.0} & \textbf{0.920} & \textbf{89.0} & \textbf{0.917} & \textbf{89.5} & \textbf{0.910} \\
    \bottomrule
  \end{tabular}
\end{table}

To demonstrate the necessity of each step in the Semantic Grounding
Module, we further evaluate SGM on an additional element-level retrieval
task. We use nine common street-view elements, such as ``bench'' and
``streetlight'', as input queries and ask the model to retrieve images
containing the corresponding visual evidence. We compute the average
Precision@$k$ at $k \in \{5,10,20\}$, as reported in
Table~\ref{tab:sgm}. The results show that CLIP+GroundingDINO improves over CLIP-only
retrieval, while adding MLLM verification further improves P@5 from
80.0\% to 88.9\% and P@20 from 77.8\% to 87.2\%. This indicates that
each step in the SGM contributes to more reliable physical evidence
confirmation. In contrast, either using CLIP alone or adding only
GroundingDINO remains limited in physical evidence grounding, especially
when visually similar street-view elements cause false positives.

\begin{table}[!ht]
  \caption{Retrieval precision of SGM stages, averaged over nine urban
    element types. Best results per column in \textbf{bold}.}
  \label{tab:sgm}
  \centering
  \small
  \setlength{\tabcolsep}{3pt}
  \begin{tabular}{lccc}
    \toprule
    Method & P@5 (\%) $\uparrow$ & P@10 (\%) $\uparrow$ & P@20 (\%) $\uparrow$ \\
    \midrule
    CLIP only              & 73.3 & 76.7 & 72.8 \\
    CLIP + GroundingDINO            & 80.0 & 80.0 & 77.8 \\
    \textbf{CLIP+GroundingDINO+MLLM}& \textbf{88.9} & \textbf{87.8} & \textbf{87.2} \\
    \bottomrule
  \end{tabular}
\end{table}

\subsection{Performance Across Query Tasks}

\begin{figure}[!t]
  \centering
  \includegraphics[width=\linewidth]{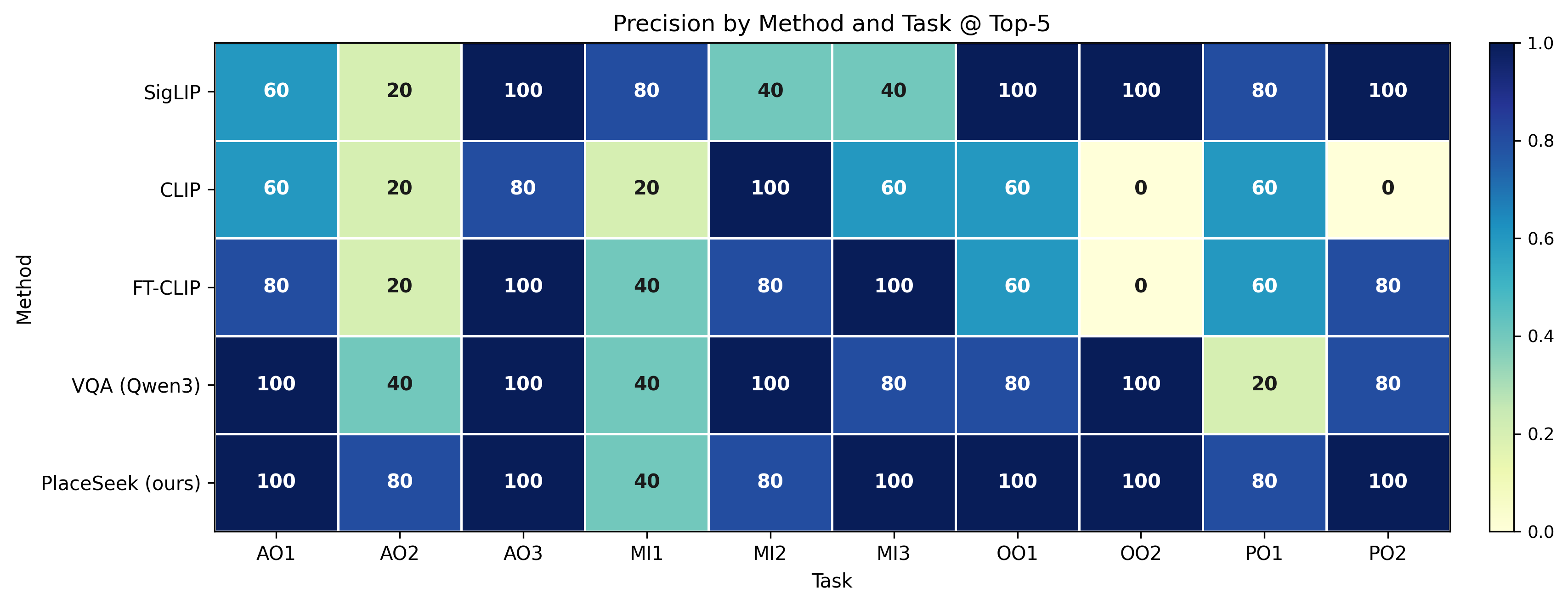}
  \caption{Per-query Precision@5 ($\uparrow$) heatmap for the five main
    methods across all 10 test queries. Darker color indicates higher
    Precision@5. PlaceSeek (bottom row) achieves perfect precision on
    six tasks and remains the most stable method overall.}
  \label{fig:heatmap}
\end{figure}

Figure~\ref{fig:heatmap} presents per-query Precision@5 for the five
main methods across all ten query tasks. PlaceSeek reaches 100\%
Precision@5 on six tasks (AO1, AO3, MI3, OO1, OO2, and PO2) and is
best or tied for best among the main methods on eight tasks. This
indicates that the full pipeline performs consistently across
activity-oriented, object-oriented, perception-oriented, and
mixed-intent queries.

In contrast, baseline performance varies substantially across query
types. CLIP and FT-CLIP perform well on queries whose target scenes
have strong overall visual cues, such as tree-lined shaded paths in AO3
or waterside urban scenes in MI2. The VQA baseline performs
competitively on several tasks, but is weaker on queries requiring
abstract affective needs (PO1, MI1) or implicit activity-support
inference (AO2). SigLIP performs well on object-oriented and
perception-oriented queries, but shows clearer limitations on
activity-oriented and mixed-intent tasks. Overall, PlaceSeek shows the
most stable profile across the full query set because it combines
explicit physical grounding with affective alignment.

The most challenging case is MI1, which requires ``a romantic public
square with a visible statue or sculpture''. Most methods fail to
achieve strong performance on this task, and PlaceSeek reaches only
40.0\% Precision@5. By reporting physical grounding and affective
alignment scores for each method-query pair,
Appendix~\ref{app:component-heatmaps} further reveals the source of this
failure: PlaceSeek obtains a physical score of 3.31 on MI1, but its
affective score is only 2.67. This suggests that
the main difficulty lies in aligning the affective concept of
``romantic'', reflecting the fact that AAM is trained on only six Place
Pulse perceptual dimensions and may not fully capture affective concepts
that deviate substantially from those dimensions.

\begin{figure*}[!t]
  \centering
  \includegraphics[width=\textwidth]{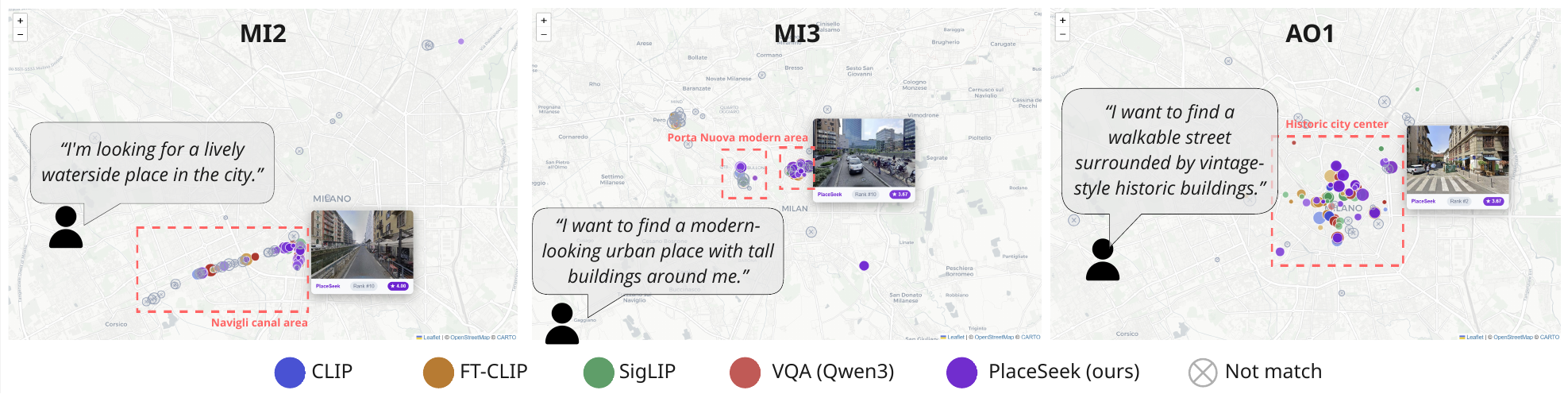}
  \caption{Geolocated top-5 retrieval results on the Milan map for three
    queries (MI2, MI3, AO1). Retrieved locations show spatial coherence with Milan's
    urban structure, including waterside, modern-area, and historic-street patterns. An interactive version covering all query tasks is
    available at \url{https://placeseek-map-production-51f3.up.railway.app}.}
  \label{fig:map-examples}
\end{figure*}

\subsection{Qualitative Analysis}
\begin{figure}[!t]
  \centering
  \includegraphics[width=\linewidth]{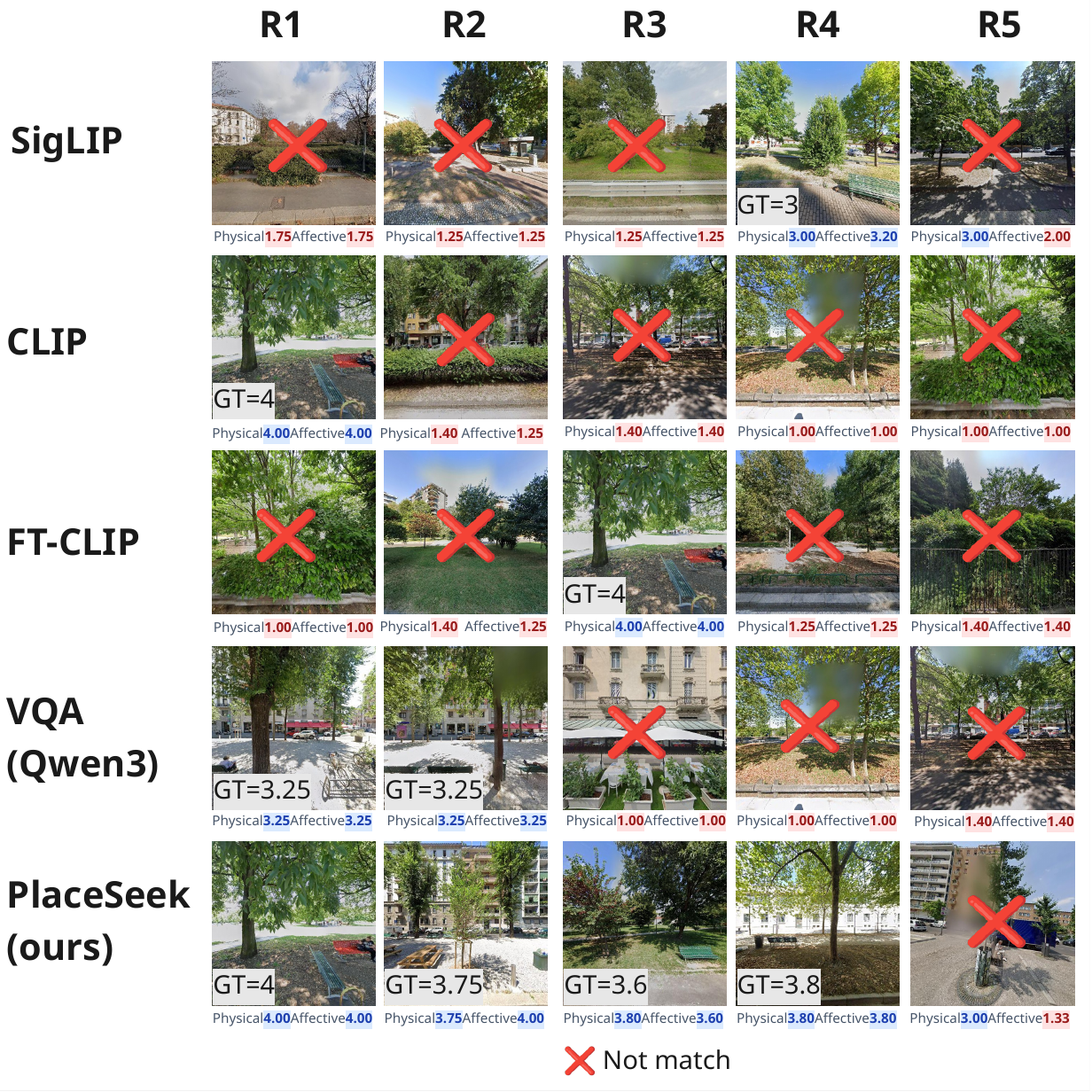}
  \caption{Top-5 retrieval results for query AO2 (a safe, quiet place with lots 
  of greenery where people can read outdoors.).}
  \label{fig:examples}
\end{figure}

Figure~\ref{fig:examples} illustrates top-5 retrieval results for query
AO2, ``a safe, quiet place with lots of greenery where people can read
outdoors.'' The baseline methods often satisfy only part of this
compositional request. SigLIP, CLIP, and FT-CLIP retrieve visually green
scenes, but many top-ranked results are not judged as good matches by
annotators. The physical and affective sub-scores indicate that these
failures are often due to missing reading-support affordances, such as
benches, and weak alignment with the expected safe and
quiet atmosphere. VQA retrieves some successful seating-related scenes,
but still includes several mismatched places.

In contrast, four of the five PlaceSeek results are successful matches
(GT $\geq 3$), combining shaded greenery with benches or other resting
affordances and a quiet public-space character. The final PlaceSeek
result is also marked as a failed match. Although it contains a seating
element, its low affective score indicates that annotators did not
perceive it as sufficiently safe or quiet for the intended outdoor
reading activity.

Figure~\ref{fig:map-examples} further visualizes geolocated top-5
retrieval results on the Milan map for three representative queries:
MI2, MI3, and AO1. The spatial distributions show that PlaceSeek
retrieves not only visually plausible images, but also geographically
coherent outdoor places. For MI2, which asks for a lively waterside
place, PlaceSeek results concentrate along the Navigli canal area, a
spatially appropriate district for waterside urban activities. For MI3,
which targets a modern-looking urban place with tall buildings, the
retrieved locations cluster around Milan's modern business districts,
including the Porta Nuova area. For AO1, which asks for a walkable
street surrounded by vintage-style historic buildings, the results are
concentrated in and around the historic city center.

In comparison, baseline methods produce more spatially scattered results
and include more failed matches, shown as grey crosses. These examples
suggest that PlaceSeek improves retrieval at both the image level and
the geospatial level: by combining physical grounding with affective
alignment, it returns candidate places that are more consistent with the
user's intended urban context and can be more directly inspected on a
map. An interactive map covering all query tasks is available at \url{https://placeseek-map-production-51f3.up.railway.app}.

\balance

\section{Conclusion and Future Work}
\label{sec:conclusion}

We formulate human-centered urban outdoor place retrieval as an intent-aware
geospatial retrieval problem and present \textit{PlaceSeek}, a framework that
maps natural-language queries to geolocated street-view results by jointly
verifying physical evidence and aligning affective expectations.
Experiments on Milan SVIs confirm that PlaceSeek outperforms
all baselines, with ablation results showing that physical grounding is
essential for retrieval validity while affective alignment refines ranking
quality among grounded candidates.

Several limitations point to future work. 
First, the evaluation is conducted in a single city, 
and covers ten query tasks with five annotators; broader studies are needed 
to test generalizability across cities, cultures and user intents. 
Second, affective alignment is constrained by the six 
perceptual dimensions in Place Pulse 2.0, which cannot fully capture richer 
and diverse affective semantics such as romantic or cozy. 
Future work should incorporate broader human perception datasets to support more 
expressive affective retrieval. Third, SVI provides only a partial representation of urban places:
visual appearance alone cannot determine whether a place is actually safe or accessible, 
which may require additional evidence such
as socioeconomic indicators, crime statistics, or POI context. Practical
deployment should further incorporate user-specific spatial constraints such
as distance, route connectivity, and nearby
facilities~\cite{yu2025spatialrag,lou2026mobility,tang2024itinera}.
PlaceSeek highlights the potential of incorporating SVI evidence and
human-centered semantics into queryable, experience-centered place
representations, pointing to a broader opportunity for next-generation
geospatial retrieval systems~\cite{mai2025nextgeoai,hou2025urbansensing}.

\begin{acks}
The authors thank the five annotators for their time and effort in
evaluating the retrieval results. Large language model tools, including
ChatGPT, were used to assist with manuscript language editing and
grammar revision.
\end{acks}

\bibliographystyle{ACM-Reference-Format}

\clearpage
\appendix
\section{Ranking Details}
\label{app:ranking-details}

\subsection{Term-Level Presence Score}
The fixed weights used in the rule-aware physical reranker are design
parameters selected from pilot inspection and kept unchanged for all
evaluation queries. For the detector-based presence score in
Section~\ref{sec:sgm}, we instantiate the normalized box-count and
box-area terms as:
\begin{align*}
  \hat{n}_{\mathrm{box}} &=
  \frac{\log(1+n_{\mathrm{box}})}{\log(1+5)},\\
  \hat{a}_{\mathrm{box}} &=
  \min\left(\frac{a_{\mathrm{box}}}{0.25},1\right),
\end{align*}
where $n_{\mathrm{box}}$ is the number of GroundingDINO boxes for the
target term and $a_{\mathrm{box}}$ is the largest detected box area ratio.
The DINO score weights are:
\[
  (\alpha_1,\alpha_2,\alpha_3)=(0.45,0.25,0.30),
\]
giving the largest weight to maximum box confidence while retaining box
count and visible scale as supporting evidence. The term-level fusion
weight is $\eta=0.65$, so the Qwen3-VL verification score receives higher
weight than the detector-only score. For not-exist rules, a forbidden
term is treated as a violation when Qwen3-VL verifies the evidence or
when the fused presence score exceeds $0.45$.

\subsection{Physical Rule Score}
For a candidate view~$v$, the physical reranker aggregates term-level
presence scores by rule type:
\begin{align*}
  s_{\mathrm{must}}(v) &= \min_{w\in\mathcal{M}} p_w(v),\\
  s_{\mathrm{more}}(v) &= \operatorname{mean}_{w\in\mathcal{M}^+} p_w(v),\\
  s_{\mathrm{less}}^{\mathrm{clean}}(v) &= 1-\operatorname{mean}_{w\in\mathcal{M}^-} p_w(v),\\
  s_{\mathrm{forbid}}^{\mathrm{clean}}(v) &= 1-\max_{w\in\mathcal{F}} p_w(v).
\end{align*}
The physical rule score is a normalized weighted sum over the components
that are active for a query:
\begin{align*}
  s_{\mathrm{phys}}(v)=\frac{1}{W(v)}(&
    0.45\,s_{\mathrm{must}}(v)
    +0.20\,s_{\mathrm{more}}(v)\\
    &+0.20\,s_{\mathrm{less}}^{\mathrm{clean}}(v)
    +0.20\,s_{\mathrm{forbid}}^{\mathrm{clean}}(v)\\
    &+0.10\,\tilde{s}_{\mathrm{clip}}(v)).
\end{align*}
Here $\mathcal{M}$, $\mathcal{M}^+$, $\mathcal{M}^-$, and $\mathcal{F}$
denote must-have, more-better, less-better, and not-exist term sets;
$\tilde{s}_{\mathrm{clip}}$ is the min-max normalized CLIP score; and
$W(v)$ is the sum of weights for the components available in that query.

\subsection{Final Physical-Affective Score}
For the final physical-affective ranking, raw affective scores
$s_{\mathrm{aff}}^{\mathrm{raw}}(v)$ are min-max normalized over the
physically verified candidate set~$\mathcal{Q}$ to obtain
$\tilde{s}_{\mathrm{aff}}(v)$. Missing affective scores are filled with
$0$ before normalization. The CLIP score $\tilde{s}_{\mathrm{clip}}(v)$
is the normalized coarse-retrieval score carried over from the physical
stage.

The fixed final fusion weights are:
\[
  (\omega_{\mathrm{phys}},\omega_{\mathrm{aff}},
  \omega_{\mathrm{clip}})=(0.65,0.25,0.10).
\]
These values keep the physically verified score as the dominant signal,
use the affective score to refine perceptual alignment, and retain CLIP
as a weak global semantic prior. After applying the physical gate, valid
candidates are sorted by
$(s_{\mathrm{final}}\downarrow,\;
s_{\mathrm{aff}}^{\mathrm{raw}}\downarrow,\;
\tilde{s}_{\mathrm{clip}}\downarrow)$.

\section{Intent Parsing Outputs}
\label{app:intent-outputs}

Table~\ref{tab:appendix-intent-outputs} summarizes the generated prompt
banks used by PlaceSeek for all ten evaluation queries. The physical
column reports the main visual-evidence families and rule types derived
from the physical prompt banks and rule files. The affective column
summarizes the perceptual prompt families used by the AAM.

\begin{table*}[t]
  \caption{Summary of generated prompt-bank outputs for all evaluation
    queries. AO2 is shown in detail in Table~\ref{tab:intent-example}.}
  \label{tab:appendix-intent-outputs}
  \centering
  \scriptsize
  \setlength{\tabcolsep}{2.5pt}
  \renewcommand{\arraystretch}{1.05}
  \begin{tabular}{p{0.06\textwidth}p{0.24\textwidth}p{0.33\textwidth}p{0.31\textwidth}}
    \toprule
    Vis. ID & Query intent & Physical prompt-bank families & Affective prompt-bank families \\
    \midrule
    AO1 &
    Walkable historic street &
    Sidewalk / pedestrian pavement (must); heritage or vintage building
    facade (must); low-traffic street (less-better); ornate architectural
    details. &
    Walkable / comfortable walking environment; vintage-style and historic
    atmosphere; charming, beautiful, and pleasant street character. \\

    AO2 &
    Safe quiet green reading place &
    Greenery (must); bench / public seating (must; inferred from reading
    outdoors). &
    Safe public place; quiet outdoor place; calm and peaceful reading
    atmosphere. \\

    AO3 &
    Tree-lined shaded jogging path &
    Tree canopy (must); shaded path; wide pavement / broad jogging path
    (must). &
    Comfortable jogging route; pleasant shaded walkway; pleasant outdoor
    setting. \\

    MI1 &
    Romantic square with statue or sculpture &
    Statue / sculpture (must); open paved square or public plaza. &
    Romantic public square; beautiful / scenic plaza; inviting and pleasant
    outdoor atmosphere. \\

    MI2 &
    Lively waterside place &
    Waterfront / river / canal / water edge (must); outdoor crowd or people
    gathering; waterside promenade or boardwalk. &
    Lively, energetic, vibrant waterside atmosphere; pleasant promenade or
    gathering place. \\

    MI3 &
    Modern place with tall buildings &
    Skyscrapers / high-rise buildings (must); modern skyline. &
    Modern streetscape; modern architecture; contemporary urban design and
    modern cityscape. \\

    OO1 &
    Well-maintained street with murals or graffiti &
    Artistic mural / graffiti (must); clean pavement; tidy sidewalk;
    intact or well-maintained building facade. &
    Well-maintained / well-kept urban space; artistic or creative street;
    vibrant, inviting, and energetic street atmosphere. \\

    OO2 &
    Tree-lined street with tram tracks &
    Tree-lined street (must); tram tracks / rails / tramway (must). &
    Charming or picturesque street scene; pleasant and comfortable urban
    setting. \\

    PO1 &
    Relaxing comfortable public square &
    Public square / city plaza (must); benches / public seating; greenery
    or plantings in square. &
    Relaxing public square; comfortable seating environment; inviting,
    welcoming, and pleasant plaza atmosphere. \\

    PO2 &
    Wealthy and safe street &
    Street and building evidence (must); high-end storefronts / luxury
    shopfronts; well-maintained sidewalk; street lighting; negative prompts
    for undesirable visual conditions. &
    Wealthy / affluent streetscape; safe and secure urban place; clean,
    tidy, and orderly street atmosphere. \\
    \bottomrule
  \end{tabular}
\end{table*}

\section{Component-Level Query Scores}
\label{app:component-heatmaps}

\begin{figure*}[t]
  \centering
  \begin{minipage}{0.49\textwidth}
    \centering
    \includegraphics[width=\linewidth]{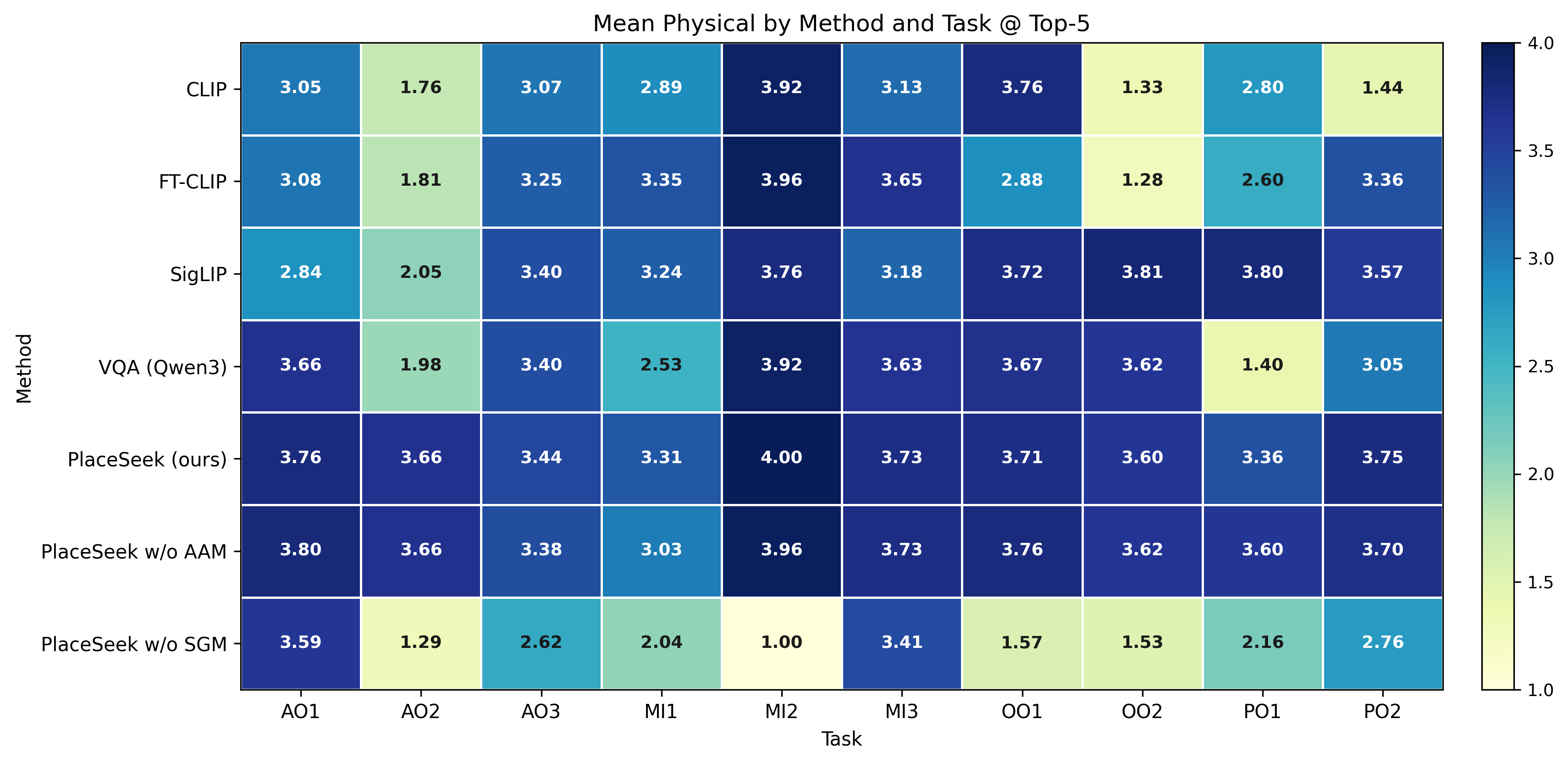}
  \end{minipage}
  \hfill
  \begin{minipage}{0.49\textwidth}
    \centering
    \includegraphics[width=\linewidth]{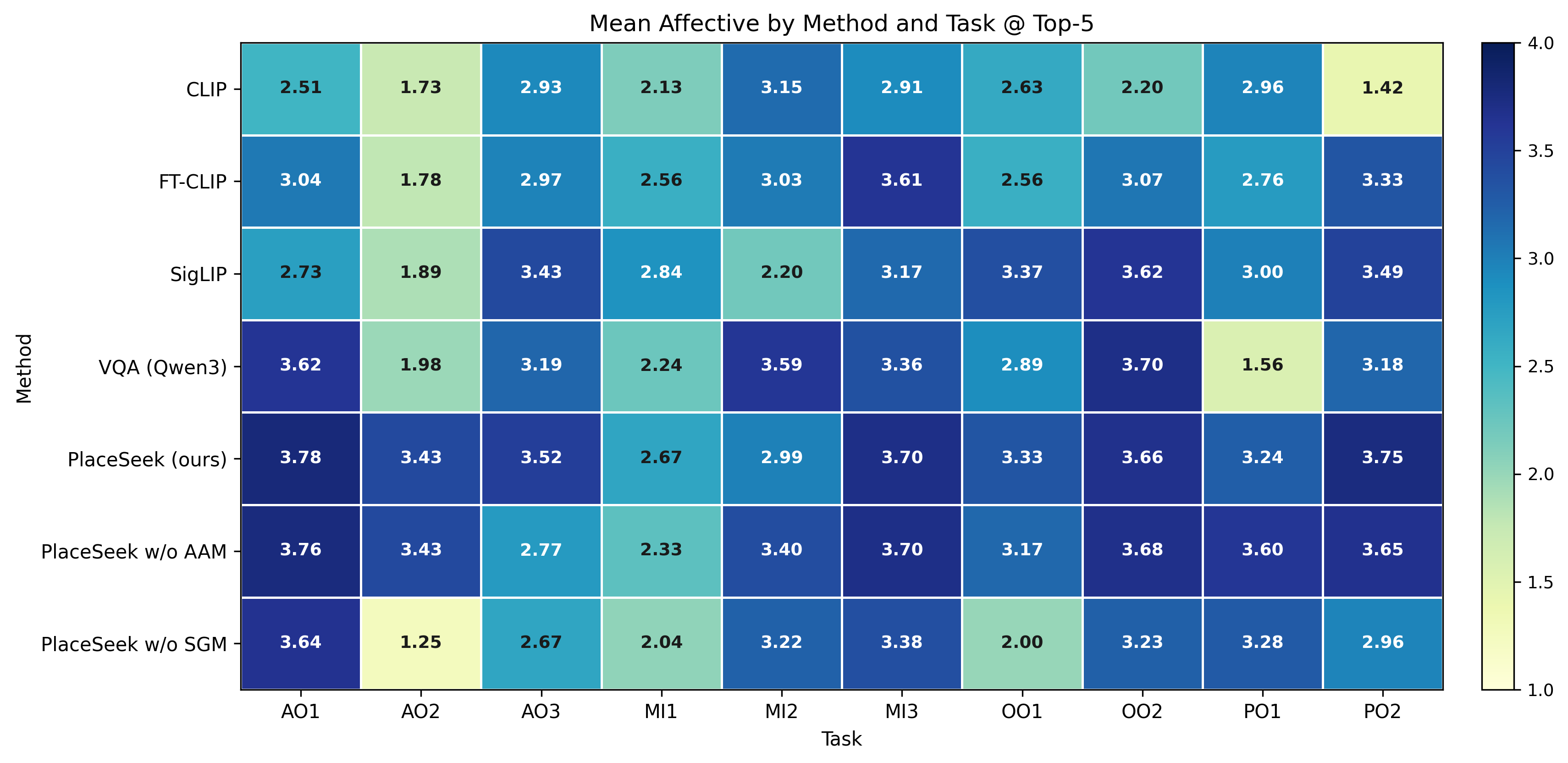}
  \end{minipage}
  \caption{Per-query top-5 component scores by method. Left: mean
    physical match score. Right: mean affective match score.}
  \label{fig:component-heatmaps}
\end{figure*}

Figure~\ref{fig:component-heatmaps} reports per-query top-5 physical and
affective match scores, providing diagnostic evidence for cases where
physical grounding and perceptual alignment diverge.

\section{Prompt Skeletons and Verification Prompt}
\label{app:prompts}

\subsection{Intent Parser Output Schema}
The full intent-parser prompt is omitted for space, but the parser output
follows the structured schema below. The values shown are illustrative
AO2-style examples for clarifying field semantics, not the complete
parser output for an actual query. This schema exposes the information
passed from LLM intent parsing to the SGM and AAM.

{\scriptsize
\begin{verbatim}
{
  "physical_requirements": [
    {
      "term": "bench / public seating",
      "source": "explicit | inferred_from_activity | inferred_from_context",
      "rule_type": "must-have | more-better | less-better | not-exist",
      "requiredness": "must | optional | forbidden",
      "visual_rationale": "why this evidence should be visible in SVI"
    }
  ],
  "affective_preferences": [
    {
      "term": "safe",
      "source": "explicit | inferred_from_context",
      "polarity": "positive | negative",
      "perceptual_family": "safety | quietness | beauty | liveliness | other",
      "rationale": "why this affective need is relevant"
    }
  ],
  "activities": ["reading outdoors"],
  "constraints": ["any query-specific tensions or exclusions"]
}
\end{verbatim}
}

\subsection{Qwen3-VL Verification Prompt}
The following prompt is used to verify cropped regions produced by
GroundingDINO. The placeholders are filled with query-specific target and
negative descriptions.

{\scriptsize
\begin{verbatim}
You are an expert urban environment analyst reviewing street-view images.

You are given a cropped image region extracted from a street-view photo.
A detection model has flagged this region as a candidate for the following type of area:

{{TARGET_DESCRIPTIONS}}

The user's intended use or requirement is:

{{INTENDED_PURPOSE}}

Your task is to independently verify whether this cropped region truly shows
one of the candidate area types above, AND whether it is genuinely suitable
for the user's intended use or requirement.
Treat the detection as a suggestion only - it may be wrong. Be a skeptical reviewer.

Return "yes" only if ALL of the following are true:
1. The area clearly matches one or more of the descriptions above.
2. The area is genuinely usable and accessible for the user's intended use.
3. Overall quality and condition appear adequate.

Return "no" if ANY of the following apply:
{{NEGATIVE_DESCRIPTIONS}}
- The region does not resemble any of the descriptions above at all.
- The area may be usable for some other purpose, but not for the user's intended use.
- The area is clearly unusable or inaccessible for the user's intended use.

Return "uncertain" if the image is too blurry, too small, heavily occluded,
or genuinely ambiguous to judge confidently.

Output JSON ONLY. No extra text outside the JSON object.

{
  "is_target": "yes",
  "confidence": 0.95,
  "reason": "brief visual evidence"
}
\end{verbatim}
}

\end{document}